\documentclass[sigconf]{acmart}
\usepackage{multirow}

\AtBeginDocument{%
  \providecommand\BibTeX{{%
    \normalfont B\kern-0.5em{\scshape i\kern-0.25em b}\kern-0.8em\TeX}}}

\copyrightyear{2024}
\acmYear{2024}
\setcopyright{acmlicensed}
\acmConference[ICMR '24] {Proceedings of the 2024 International Conference on Multimedia Retrieval}{June 10--14, 2024}{Phuket, Thailand.}
\acmBooktitle{Proceedings of the 2024 International Conference on Multimedia Retrieval (ICMR '24), June 10--14, 2024, Phuket, Thailand}
\acmPrice{15.00}
\acmDOI{10.1145/3652583.3658040}
\acmISBN{979-8-4007-0619-6/24/06}
\usepackage{stfloats}
\usepackage{setspace}
\usepackage{makecell}

\begin{document}
\begin{sloppypar}

\title{G-SAP: Graph-based Structure-Aware Prompt Learning over Heterogeneous Knowledge for Commonsense Question Answering}

\author{Ruiting Dai}
\email{rtdai@uestc.edu.cn}
\orcid{1234-5678-9012}
\affiliation{%
  \institution{University of Electronic Science and Technology of China}
  \city{ChengDu}
  \country{China}
}

\author{Yuqiao Tan}
\email{yuqiaot@std.uestc.edu.cn}
\orcid{0009-0009-0693-5230}
\affiliation{%
  \institution{University of Electronic Science and Technology of China}
  \city{ChengDu}
  \country{China}
}

\author{Lisi Mo}
\email{morris@uestc.edu.cn}
\authornote{*Corresponding authors}
\affiliation{%
  \institution{University of Electronic Science and Technology of China}
  \city{ChengDu}
  \country{China}
}

\author{Shuang Liang}
\email{shuangliang@uestc.edu.cn}
\affiliation{%
  \institution{University of Electronic Science and Technology of China}
  \city{ChengDu}
  \country{China}
}

\author{Guohao Huo}
\email{gh.huo513@gmail.com}
\affiliation{%
  \institution{University of Electronic Science and Technology of China}
  \city{ChengDu}
  \country{China}
}

\author{Jiayi Luo}
\email{jiayi.luo59@gmail.com}
\affiliation{%
  \institution{University of Electronic Science and Technology of China}
  \city{ChengDu}
  \country{China}
}
\author{Yao Cheng}
\email{yaoyaoverycool@gmail.com}
\affiliation{%
  \institution{University of Electronic Science and Technology of China}
  \city{ChengDu}
  \country{China}
}

\renewcommand{\shortauthors}{Ruiting Dai et al.}

\begin{abstract} 
Commonsense question answering has demonstrated considerable potential across various applications like assistants and social robots. Although fully fine-tuned Pre-trained Language Model(PLM) has achieved remarkable performance in commonsense reasoning, their tendency to excessively prioritize textual information hampers the precise transfer of structural knowledge and undermines interpretability. Some studies have explored combining Language Models (LM) with Knowledge Graphs (KGs) by coarsely fusing the two modalities to perform Graph Neural Network (GNN)-based reasoning that lacks a profound interaction between heterogeneous modalities.
In this paper, we propose a novel \underline{\textbf{G}}raph-based \underline{\textbf{S}}tructure-\underline{\textbf{A}}ware \underline{\textbf{P}}rompt Learning Model for commonsense reasoning, named \textbf{G-SAP}, aiming to maintain a balance between heterogeneous knowledge and enhance the cross-modal interaction within the LM+GNNs model. In particular, an evidence graph is constructed by integrating multiple knowledge sources, i.e. ConceptNet, Wikipedia, and Cambridge Dictionary to boost the performance. Afterward, a structure-aware frozen PLM is employed to fully incorporate the structured and textual information from the evidence graph, where the generation of prompts is driven by graph entities and relations. Finally, a heterogeneous message-passing reasoning module is used to facilitate deep interaction of knowledge between the LM and graph-based networks. Empirical validation, conducted through extensive experiments on three benchmark datasets, demonstrates the notable performance of the proposed model. The results reveal a significant advancement over the existing models, especially, with $6.12$\% improvement over the SoTA LM+GNNs model ~\cite{huang2023mvp} on the OpenbookQA dataset. 
\end{abstract}



\maketitle

\section{Introduction}
Commonsense Question Answering (CSQA) aims to replicate human-like understanding in machines by leveraging commonsense knowledge to answer natural language questions. This field faces significant challenges in harnessing various types of commonsense knowledge, such as generally accepted rules and findings ~\cite{talmor2019commonsenseqa}. Although pre-trained language models (PLMs) ~\cite{lewis2020bart,radford2019language,yang2019xlnet,brown2020language} like BERT ~\cite{devlin2018bert} excel in tasks like entity recognition ~\cite{kenton2019bert} and sentiment analysis ~\cite{kenton2019bert}, they falter with commonsense-based queries—a domain where humans excel intuitively.

\begin{figure}[!t]
    \centering
    \includegraphics [width=\columnwidth]{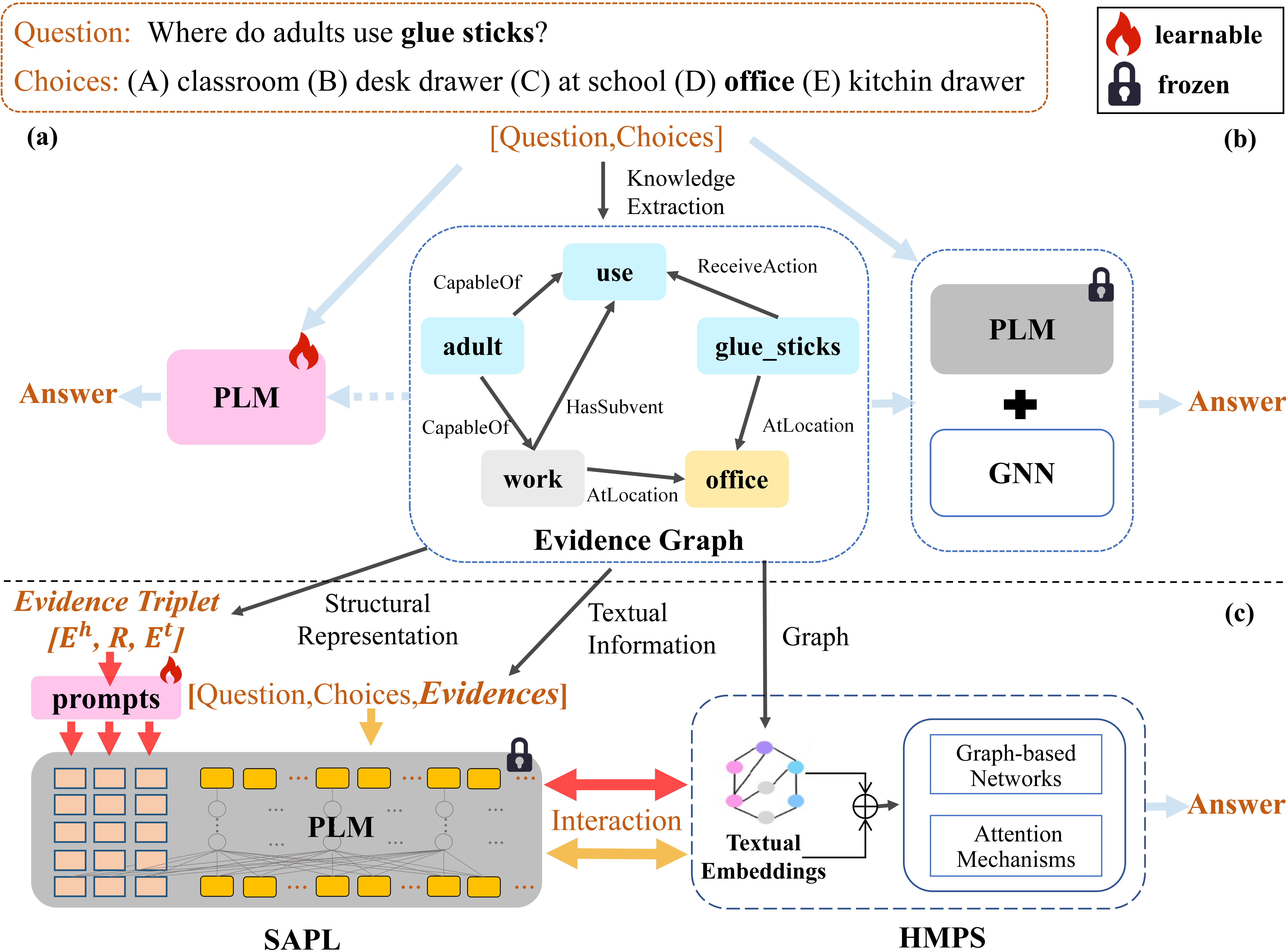}
    \caption{Comparison of existing methods and our G-SAP for CSQA. (a) PLM-based methods feed QA content, or together with the retrieved knowledge into LM for full fine-tuning. (b) LM+GNNs methods leverage shallow cross-modal interaction operations to fuse the encoded representations from LM and GNN. (c) Our G-SAP uses structure-aware prompts to fully interact KG structure information with textual information in PLM (SAPL). The outputs are then fed into HMPS to update the KG graph and textual representation for heterogeneous message-passing reasoning. 
    }
    \label{fig:first}
\end{figure}

As shown in Figure \ref{fig:first} (a), some CSQA approaches predominantly center around the direct fine-tuning of PLMs that come equipped with large-scale parameters and linguistic knowledge acquired during the pre-training phase. For instance, ~\cite{lourie2021unicorn} presents a general commonsense reasoning model that achieves exceptional performance on eight commonsense benchmark tests through fully fine-tuning the PLM Unicorn. Nonetheless, these methods either rely on pure QA textual context or incorporate triple evidence text as complementary inputs, disregarding the importance of structural knowledge. As a result, PLM-based models treat these inputs indiscriminately, leading to excessive overfitting of textual information ~\cite{geva2019we,poliak2018hypothesis}, while also hindering the precise transfer of structured knowledge and compromising interpretability ~\cite{houlsby2019parameter,lin2019kagnet}. To surmount this, recent studies ~\cite{lin2019kagnet,feng2020scalable,sun2022jointlk,zheng2022dynamic,DBLP:conf/eccv/HeGSL22, DBLP:journals/tnn/HeGSL23} couple the PLMs with knowledge graphs have been explored to bolster both accuracy and interpretability of reasoning. As depicted in Figure \ref{fig:first} (b), these approaches typically construct evidence subgraphs related to the extracted entities in QAs and subsequently employ graph neural networks (GNNs) to facilitate joint reasoning alongside the LMs.

While the combination of LMs and GNNs has shown promise in addressing CSQA tasks, several challenges remain. Some studies ~\cite{lin2019kagnet,lv2020graph,yasunaga2022deep,zheng2022dynamic} encode the QA context and KG subgraph in isolation, with a shallow fusion at the GNN layer through message passing ~\cite{yasunaga2021qa,zheng2022dynamic}, attention mechanisms ~\cite{sun2022jointlk} or at the output layer through attention mechanism ~\cite{feng2020scalable,yasunaga2022deep} or MLP ~\cite{huang2023mvp}. Other works ~\cite{wang2023dynamic,bian2021benchmarking,xu2021fusing} have endeavored to improve the interaction and integration of heterogeneous modalities by directly feeding retrieved knowledge graph entities alongside the QA context to LMs. However, these approaches exhibit a coarse-grained level of fusion that inadequately enables the interaction of heterogeneous modalities, failing to effectively address the inherent training bias between structural and textual information. This raises the question:
\textit{How can we effectively balance the treatment of structured and textual information in LM+GNNs to promote a profound interaction of heterogeneous knowledge?}

In response to these challenges, we present a novel \underline{\textbf{G}}raph-based \underline{\textbf{S}}tructure-\underline{\textbf{A}}ware \underline{\textbf{P}}rompt Learning Framework for effective commonsense reasoning, dubbed \textbf{G-SAP}. Drawing inspiration from the parameter efficiency of prompt learning, G-SAP's cornerstone lies in the structure-aware prompts that are prepended to the inputs of frozen PLMs along with the text embeddings.
Unlike naive prompts ~\cite{li2021prefix,lester2021power} which cannot represent structural information from KGs, we harness structural information from knowledge graphs to generate prompts, facilitating a balanced interaction of heterogeneous
modalities and avoiding biased training during the encoding process. 
Specifically, G-SAP extracts evidence from multiple sources, i.e. ConceptNet, Wikipedia, and Cambridge Dictionary to construct a comprehensive evidence graph. As shown in Figure \ref{fig:first} (c), we propose the structure-aware prompt vectors constructed based on the KG entities and relations representations, feeding them into the frozen PLMs to fuse the textual and structure knowledge. The fused prompts and textual embeddings are used as inputs to the heterogeneous message-passing reasoning module (HMPS) that iteratively updates the representation of the evidence graph and produces the final answer prediction. Furthermore, we design graph-based networks and attention mechanisms to respectively fuse and reason over the heterogeneous modalities in HMPS. In summary, the contributions of this paper are threefold as follows:
\begin{itemize}
    \item To the best of our knowledge, we are the first to propose a graph-based structure-aware prompt learning model for commonsense reasoning, where prompts are generated based on graph-structural representations.

    \item We present a novel heterogeneous message-passing reasoning strategy that leverages graph-based networks and attention mechanisms to fuse and reason over both textual and structural information, thereby deepening the interaction between heterogeneous knowledge.

    \item We conduct extensive experiments on three CSQA benchmark datasets across a number of evaluation metrics, ablation studies, and case studies. Our results significantly outperform current state-of-the-art models (e.g., with $6.12$\% improvements over the SoTA LM+GNNs model ~\cite{huang2023mvp} on the OpenbookQA dataset). 
\end{itemize}

\section{Related Work}

\subsection{Graph-based methods for CSQA}
Graph-based methods for CSQA encompass two main categories: Relation Network-based (RN-based) models and Graph Neural Network (GNN)-based models. RN-based models effectively leverage knowledge by modeling the relationship paths within the knowledge graph. For instance, KagNet ~\cite{lin2019kagnet} enhances the capabilities of the Relation Network (RN) ~\cite{santoro2017simple} by incorporating multi-hop path modeling. 
On the other hand, GNN-based models offer improved scalability by employing a message-passing mechanism. An example is the GNN-based model proposed by ~\cite{lv2020graph} for CSQA that integrates heterogeneous knowledge sources. It construct graphs from structured knowledge bases and textual information, enabling graph-based reasoning. Additionally, ~\cite{feng2020scalable} presents a multi-hop graph relation network (MHGRN) combines the scalability of graph networks with the interpretability of path-based models.

Graph-based methods often necessitate domain expertise to design GNN modules that effectively encode knowledge subgraphs. These methods typically concentrate solely on structured knowledge associated with entities, disregarding the potential inclusion of text-based knowledge that may be relevant to the query.

\subsection{PLM-based methods for CSQA}

The increasing adoption of Transformer-based PLMs~\cite{han2021transformer} such as BERT ~\cite{kenton2019bert}, GPT ~\cite{Radford2018ImprovingLU}, and XLNet \cite{yang2019xlnet} in natural language processing has led to a growing interest in transferring these large-scale models to the domain of CSQA. Initial studies ~\cite{lourie2021unicorn,khashabi2020unifiedqa,lourie2021unicorn,khashabi2020unifiedqa} have focused on straightforward transfer and parameter fine-tuning of these pre-trained models. Recent studies propose combining language models with knowledge graphs to leverage their strengths ~\cite{lin2019kagnet,yasunaga2021qa}, with LMs capturing implicit patterns while KGs explicitly represent structural relations. Some studies ~\cite{wang2023dynamic,bian2021benchmarking,xu2021fusing} aim to enhance interaction and integration of diverse information by directly feeding retrieved entities from knowledge graphs with question answering contexts for LMs. However, these approaches lack discriminative and fine-grained fusion, resulting in imbalance between handling knowledge graphs and text-based QA. Other approaches ~\cite{lin2019kagnet,lv2020graph,yasunaga2022deep,zheng2022dynamic} encode QA context and KG subgraph separately and only perform shallow fusion at GNN layer using techniques like message passing or attention, limiting reasoning capability.

In this paper, we introduce a deep interaction approach that effectively integrates the QA context and subgraph knowledge within the PLM module and the message passing module, aiming to enhance commonsense question answering.

\subsection{Prompt Learning}
While PLM-based methods have achieved notable results, challenges persist in the fine-tuning process, particularly due to the escalating computational costs. The seminal work ~\cite{brown2020language} revealed the efficacy of manually designed text templates, referred to as prompts, in the GPT3 ~\cite{brown2020language}. Prompt engineering, which involves combining discrete or continuous prompts with Masked Language Modeling  (MLM) ~\cite{kim2021vilt}, serves to alleviate the computational burden. However, early manual prompt engineering incurs substantial human costs and presents difficulties in end-to-end optimization. Recent studies like ~\cite{li2021prefix,lester2021power,yong2023prompt} relax the constraints of discrete templates by employing trainable continuous vectors within frozen PLMs. These PLMs with fewer parameters have achieve comparable performance across various natural language processing (NLP) tasks. 

Inspired by the above, we are the first to propose the integration of graph-based message-passing reasoning with structure-aware prompt learning for commonsense reasoning.

\section{Method}

\begin{figure*} [t]
    \centering
    \includegraphics [width=0.98\textwidth]{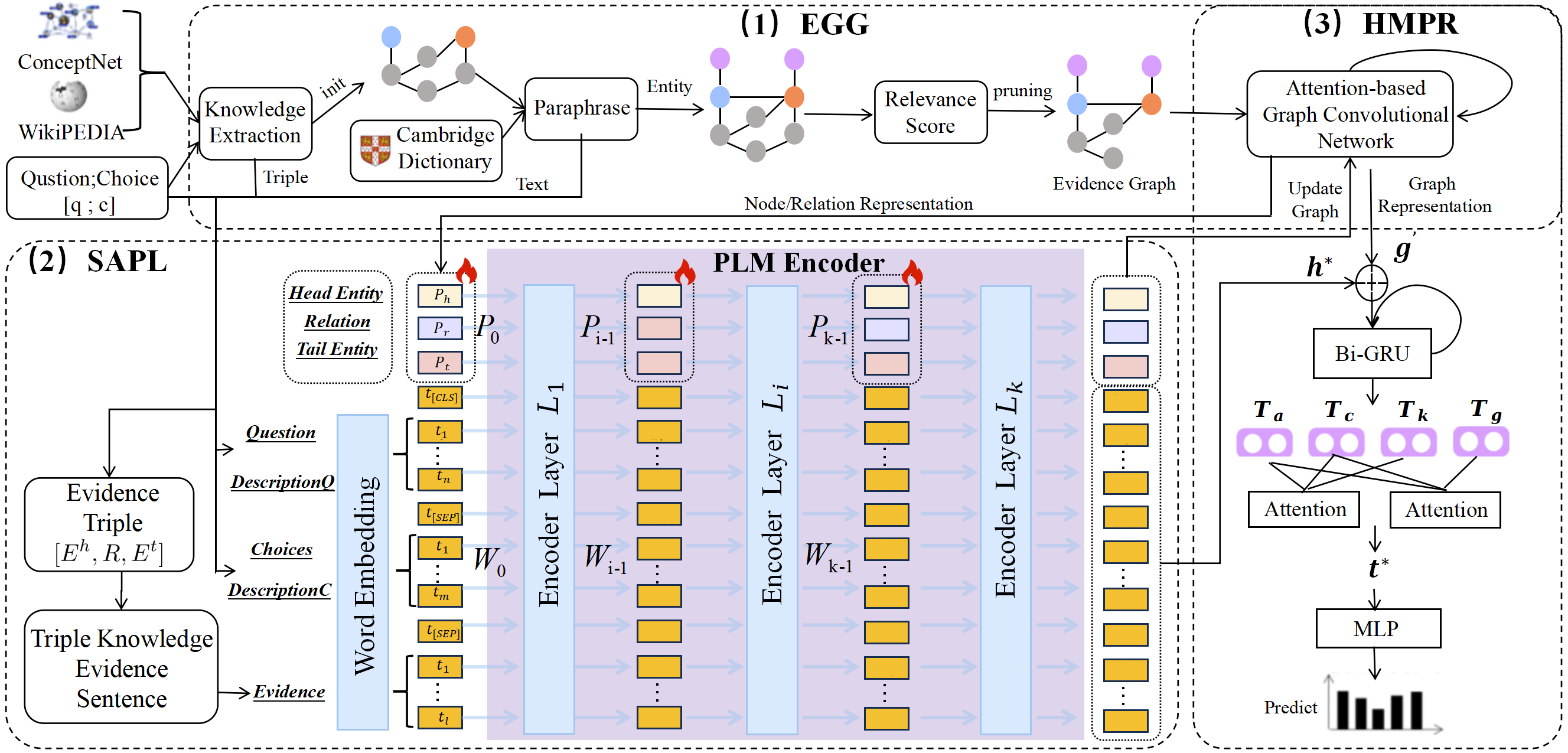}
    \caption{Overall framework of G-SAP. We first extract evidence from sources to obtain an emerged evidence graph, then introduce paraphrase and relevance scores to refine it. Secondly, structure-aware prompts are constructed based on KG nodes and relations representations, enabling the inclusion of both structural and textual information in the frozen PLMs. Finally, a heterogeneous messaging-passing strategy is employed, utilizing graph-based attention networks that fuse the PLM results to iteratively update the evidence graph representation and generate the final answer prediction.}
    
    \label{fig:enter-label}
\end{figure*}

\subsection{Task Definition}\label{AA}
This paper focuses primarily on the task of CommonSense Question Answering (CSQA) in a multiple-choice format. Given a natural language question q and a set of candidate choices $\{c_1, c_2, ..., c_b\}$, our objective is to accurately distinguish the correct answer from the incorrect ones, with accuracy serving as the evaluation metric. Initially, we extract entities using KeyBERT~\cite{grootendorst2020keybert} and preprocessing techniques, denoted as $V_{topic}=\{V_q \cup V_c\}$, where $V_q$ represents the set of entities in questions and $V_c$ represents the set of entities in choices. Subsequently,  our work involves extracting evidence from relevant commonsense knowledge bases. Due to the scarcity of knowledge and the complexity of network parameters, we simplify the approach by treating all edges within the KGs as undirected.

\subsection{Overall Framework}\label{BB}

As illustrated in Figure \ref{fig:enter-label}, \textbf{G-SAP} encompasses three essential modules: an evidence graph generation (\textbf{EGG}) module, a structure-aware prompt learning (\textbf{SAPL}) module, and a heterogeneous message passing reasoning (\textbf{HMPR}) module. Specifically, the evidence graph generation module aims to enrich the evidence graph by incorporating external knowledge from Concept, Wikipedia, and Cambridge Dictionary, which plays a crucial role in knowledge reasoning tasks. The structure-aware prompt learning module is trained using both structural knowledge and textual information to achieve a balanced fusion of heterogeneous knowledge, thereby mitigating the over-fitting issue of textual information. To facilitate a deep interaction of heterogeneous knowledge, we propose a heterogeneous message-passing reasoning module that utilizes graph-based networks and attention mechanisms to respectively fuse and reason over the different modalities. This enables the reasoning networks to concurrently assimilate both structural- and textual-level features.
\subsubsection{Evidence Graph Generation Module}

As background knowledge is not provided in CSQA tasks, it is necessary to retrieve evidence subgraphs from the knowledge graph to conduct answer reasoning. Therefore, similar to ~\cite{lv2020graph}, we retrieve evidence from ConceptNet and Wikipedia plain text to initialize evidence subgraphs, denoted as $G_{sub}=(V_{sub}, E_{sub})$, where $V_{sub}=V_{topic} \cup V_o$, $V_o$ represents the set of other evidence entities, $E_{sub}$ is the set of their connection edges. 

\noindent \textbf{Evidence Graph Generation.} To enhance the knowledge within the evidence subgraph, we also integrate entity paraphrases obtained from the Cambridge Dictionary. Unlike conventional methods such as ~\cite{lv2020graph}, which solely involve inputting the text of paraphrases into the PLMs, our method also emphasizes integrating the structure information of the paraphrase by attaching their entities to the evidence graph. Specifically, we first align and merge the subgraphs to generate a complete evidence graph, defined as $G_{pat}=(V_{pat}, 
E_{pat})$. We retrieve the noun paraphrase for $V_{topic}$ from the Cambridge Dictionary and identify their entities, representing as $V_h$. We establish connections between $V_h$ and their corresponding topic entities in $V_{topic}$, initializing the edge relationship $r_{th}$ as "DefTop". Additionally, we also connect $V_q$ and $V_c$ within $V_{topic}$, initializing their relationship $r_{qc}$ as "RelatedQA". 
Therefore, by updating the evidence graph $G_{pat}$, the node set  $V_{pat}=V_{topic} \cup V_o \cup V_h$, the edge set $E_{pat}$ is represented as $E_{pat}=E_{sub} \cup E_h$, where $E_h$ is formalized as: $E_h=\{{(\nu_q,r_{qc},\nu_c)|\nu_q\in V_q}\cup{(\nu_t,r_{th},\nu_h)|\nu_t\in V_{topic}}\}$.

\noindent \textbf{Graph Pruning and Representation}
The importance of a high-quality knowledge graph and its representation has been extensively demonstrated in facilitating subsequent inference. To further enhance the quality of the evidence graph obtained previously and the cross-modal interaction, we employ pruning operations whereby the relevance score is calculated between entities and questions, removing nodes with relevance score $\lambda$ of less than $0.1$.
For graph representation, traditional path-based methods encounter scalability issues due to the polynomial and exponential growth of the number of nodes and edges as the graph size increases. In contrast, graph neural network methods solely aggregate information from neighboring nodes, neglecting the influence of edge types. Additionally, these methods heavily rely on implicit encodings through the network, resulting in a lack of transparency and interpretability. To overcome these limitations, we propose an attention-based graph convolutional network that integrates additional edge-type features and calculates attention weights between nodes to extend the message-passing encoding mechanism.

Specifically, we account for the influences of entity, entity type, relationship type, and entity relevance within the graph. To begin, we define the node type information as $\overrightarrow{v}_{u}^{type}=f_{v}(v_{u}^{type})$, the relationship type information as $\overrightarrow{r}_{un}^{type}=f_{r}(v_{u}^{type};v_{n}^{type};r_{un})$, and the relevance score $\lambda_u = \sigma(f_{\lambda}(\phi(v_u);\phi(v_q))$, where $\overrightarrow{v}_u^{type}$, $\overrightarrow{r}_{un}^{type}$ separately represent the type of node $u$ and the relation between node $u$ and $n$ in the evidence graph in vectorization, while $v_u^{type}$ is a one-hot vector representing various classes, $f_v$, $f_r$, $f_\lambda$ are linear layers, $\lambda_u \in (0,1)$ represents the relevance of entity $u$ to the question $q$, $\phi$ is the pre-trained encoding function. As a result, the message from node $u$ to $n$ can be represented as:

\begin{equation}
m_{un}=f_{u\rightarrow n}(h_{u}^{(l-1)};\overrightarrow{v}_{u}^{type};\overrightarrow{r}_{un}^{type})
\end{equation}
where $h^{(l-1)}_u$ represents the hidden layer encoding of node $u$ at the $(l-1)$th layer, $f_{u\rightarrow n}$ is a linear layer. 

Secondly, considering that messages from various neighboring nodes may hold varying degrees of significance, we compute attention weights $a$ related to all neighbor nodes based on their node type, relationship type, and topic relevance score. This allows us to aggregate high-quality information that is most relevant to the contextual context. Taking the message from node $u$ to node $n$ as an example, the method for calculating attention weights is defined as follows:

\begin{equation}
q_{u}=f_{q}(h_{u}^{(l-1)};\overrightarrow{v}_{u}^{type};f_{\lambda}(\lambda_{u}))
\end{equation}

\begin{equation}
k_{n}=f_{k}\left(h_{n}^{(l-1)};\overrightarrow{v}_{n}^{type};f_{\lambda}\left(\lambda_{n}\right);\overrightarrow{r}_{un}^{type}\right)
\end{equation}

\begin{equation}
\alpha_{un}=softmax(\frac{q_{u}^{\top}k_{n}}{\sqrt{D}})
\end{equation}
here, $f_q$, $f_{\lambda}$, and $f_k$ refer to linear layers specifically designed to transform the feature vector dimension to a consistent size, facilitating efficient feature fusion. $D$ represents the dimension of the features.
As a result, the enhanced message-passing mechanism can be defined as follow:
\begin{equation}
h_{n}^{(l)}=f_{n}\left(\sum_{u\epsilon N_{n}\cup n}\alpha_{un}m_{un}\right)+h_{n}^{(l-1)}
\end{equation}
where $N_{n}$ represents the neighbor nodes of $n$, and $f_n$ introduces batch normalization to maintain the same distribution between each layer of the network through dimension-preserving mapping. Through the updates from the layer-wise graph network, all nodes fully incorporate the contextual information from the question as well as the structural information from the graph. Finally, we obtain the representation $g$ of the evidence graph by encoding all nodes and performing average pooling, as defined by (6).
\begin{equation}
g=\text{MeanPooling}(\{h_{\nu}^{(L)}|\nu\in V_{pat}\})
\end{equation}

\subsubsection{Structure-aware Prompt Learning module}

Although several studies ~\cite{zhou2022learning} have explored the utilization of both structural and textual information in NLP tasks, none of them can address the over-fitting issue of textual information in LM-based commonsense reasoning models. Inspired by the effectiveness of applying prompts to solve overfitting issues in frozen PLMs ~\cite{wang2022promda}, we explore the application of structure-aware prompts in PLM to avoid CSQA models overly focusing on textual information. For this module, we introduce the details of the pre-trained language model.

\noindent \textbf{Pre-trained Language Model} Assuming the pre-trained language model $\mathcal{P}$ has $L$ transformer layers and a hidden size of $H$. A prompt is a series of trainable embedding vectors added to the input of the frozen pre-trained language model. ~\cite{li2021prefix} proposed a Layerwise Prompt, which inserts relatively short prompt sequences at each layer (e.g., 5-10 vectors), allowing frequent interaction with text information about entities and relationships in the PLM. Inspired by this, we propose a novel structured prompt, with k trainable vectors at each layer. Therefore, the input of $\text{j-th}$ layer: $h^{(j)} \in R^H$ is defined as follow:\begin{equation}
\left.h^{(j)}=\left\{\begin{array}{cl}  [s^{(j)}_0,\cdots,s^{(j)}_k;h^{(j)}_{k+1},\cdots,h^{(j)}_{H}]&\quad 1 \textless j\leq p\\ \\\boldsymbol{PLM}(h^{(j-1)})&\quad j \textgreater p\end{array}\right.\right.
\end{equation}
where $\boldsymbol{PLM}(\text{·})$ is the forward function of the Transformer layer, $k$ shows the length of the prompt and $p$ defines the length of prompting layers,  $s^{(j)}_i$ is the $\text{i-th}$ prompt vector of the $\text{j-th}$ layer. Note that $h^{(1)}$ is the result of embedding functions that preprocess the inputs into token sequences. $\boldsymbol{PLM}(\text{·})$ operates on the entire sequence (the concatenation of prompt and text). 

More specifically, we can divide the input of PLM into two types: structure-aware prompts and textual embeddings. The parameters of a structure-aware prompt are generated from the evidence graph's encoding $g$, while textual embeddings are generated from information such as the question, choices, and evidence. The corresponding output would serve as the input of the graph-based reasoning module, for updating the evidence graph and reasoning the final results. This allows structural knowledge from the evidence knowledge graph and textual knowledge from the PLM to be balanced integrated through structure-aware prompt learning.

\noindent \textbf{Structure-aware Prompt} To reduce computational complexity, we primarily focus on the triplets $[E^h, R, E^t]$ connected by question entities to choice entities. We use the embeddings of their entities and relations $E^h$, $R$, $E^t$ to generate structured-aware prompt $\boldsymbol{p}_{tp} =\{p_i\in\mathbb{R}^{d}\}_{i=1}^{tp_\_count}$, $tp_\_count$ is the count of triplets. Formally,
\begin{equation}
p_i = [F(E_i^h, g);F(R_i, g);F(E_i^t, g)]
\end{equation}
\begin{equation}
F(e, g) = W_{out} \cdot (\text{ReLU}(W_{in}\cdot(e+W_g\cdot g)))
\end{equation}
where $W_{in}\in R^{d_h\times d}$ and $W_{out}\in R^{d\times d_h}$ are trainable weight matrices, where $d_h$ is the intermediate hidden size of the mapping layer and $d$ is transformer block's hidden size. Then, we reorganize $F(E^h_i,g)$, $F(R_i,g)$, and $F(E^t_i,g)$ into a series of input embeddings, evenly distributed across each PLM layer. In this way, combining with the structured embeddings obtained from the graph neural network can better enhance the effectiveness of the clues. During this process, there is comprehensive interaction between the input tokens of $\mathcal{P}$ and the structured prompt $\mathcal{S}$, 
$\mathcal{G}$ (linearly mapped to $\mathcal{S}$) and textual knowledge from $\mathcal{P}$ to be fully integrated. This balances and integrates the structural knowledge from $\mathcal{G}$ and textual knowledge from $\mathcal{P}$ through $\mathcal{S}$.

\noindent \textbf{Textual embedding} To represent a QA query, we extract and concatenate the raw tokens of the question, choices, paraphrases, and evidence, including their respective descriptions. In particular, the evidence descriptions are obtained by transforming triplets into natural language. Following BERT ~\cite{kenton2019bert}, we use a special token [SEP] to connect all the texts and feed the concatenated text into the frozen PLM $\mathcal{P}$. Formally, the concatenated text can be represented as follow:

\begin{equation}
[ \text{[CLS]}, t_q, t^q_d, \text{[SEP]}, t_c, t^c_d, \text{[SEP]},t_e]
\end{equation}
where $t_q$, $t^q_d$, $t_c$, $t^c_d$, $t_e$ represent the texts of question, question paraphrase, choices, choice paraphrases, and evidence respectively.

\subsubsection{Heterogeneous Message Passing Reasoning Module}
While some prior works ~\cite{wang2023dynamic,DBLP:conf/coling/ChenJCZ20} have explored various graph-based knowledge reasoning strategies for CSQA tasks, they often adopt a coarse fusion approach, overlooking the different influences of textual and structure information, resulting in unexplainable reasoning. To address this limitation, our proposed approach incorporates a bidirectional recurrent neural network for coarse fusion of two modalities: textual and structural knowledge, along with an attention network for deeper knowledge cross-modal interactions.

Specifically, we incorporate the output of structure-aware prompts $\boldsymbol{p}_{tp}$ and textual embeddings from PLM into the graph-based reasoning module. $\boldsymbol{p}_{tp}$ is for enhancing the representation of evidence graph, and textual embedding is for joint reasoning. We extract all stripe's embeddings from $\boldsymbol{p}_{tp}$, and each stripe embedding can be divided into [$E^h$, $R$, $E^t$]. Therefore, the entity representation in the evidence graph can be updated by corresponding $E$. Notably, the representation of the question entity is calculated by summing and normalizing multiple representations that occur in multiple relations to different choices. Hence, we proceed to enhance the evidence graph representation, denoted as $g^\prime$, by employing the attention-based graph convolutional network as described in section 3.2.1.
Moreover, we also calculate the textual embedding by summing and normalizing multiple textual representations and denote it as $h^\ast$. 

To initially fuse cross-modal knowledge, We concatenate  $h^\ast$ with $g^\prime$ and feed them into a bidirectional recurrent neural network as follows:
\begin{align}
\overrightarrow{x_t} &= f_{fw}([h^\ast;g'], \overrightarrow{x}_{t-1}) , \;
\overleftarrow{x_t} = f_{bw}([h^\ast;g'], \overleftarrow{x}_{t+1})
\end{align}
\begin{equation}
T_{context} = \boldsymbol{W}\cdot f_{contact}([\overrightarrow{x}_T, \overleftarrow{x}_1]) + b
\end{equation}
\\
where $\overrightarrow{x_t}$ and $\overleftarrow{x_t}$ represent the forward and backward computation results of the bidirectional GRU~\cite{cho2014learning}, obtained by $f_{fw}$ and $f_{bw}$. Finally, $\overrightarrow{x_t}$ and $\overleftarrow{x_t}$ are concatenated and sent into a layer of MLP to obtain $T_{context}$.
According to the fusion information,  we divide $T_{context}$ into four groups as $T_{context}=[T_{a}; T_{c}; T_{k}; T_{g}]$, representing question and choice, evidence, and evidence graph knowledge respectively. Simultaneously, we employ the knowledge attention method to calculate the contextual attention scores for unstructured knowledge and structured knowledge separately respectively. As the attention score $\alpha_{ka}$ of $T_k$ and $T_a$ is computed in (\ref{attn}), we get $\alpha_{kc}$ of $T_k$ and $T_c$, $\alpha_{ga}$ of $T_g$ and $T_a$, $\alpha_{gc}$ of $T_g$ and $T_c$ similarly. 
\begin{equation}
\label{attn}
\alpha_{ka} = \text{attention}(T_k, T_a)
\end{equation}
Next, we obtain the deep fusion embedding of heterogeneous knowledge as follow:
\begin{equation}
\label{12}
t^{\ast}=[T_{a};T_{c};(\alpha_{ka}+\alpha_{kc})T_{k};(\alpha_{ga+}\alpha_{gc})T_{g}]
\end{equation}
Finally, we employ a linear classification layer to generate a set of deeply fused feature vectors as follow:
\begin{equation}
o=\text{ReLU}(\boldsymbol{W}_h t^\ast)=\{o_i|i=1,...,5\}
\end{equation}
where $t^\ast$ represents the fused representation. The predicted result is calculated by identifying the label corresponding to the maximum confidence score as:
\begin{equation}
\text{pred}=\text{label}(\text{argmax}(o))
\end{equation}
this approach ensures that the model prioritizes knowledge information directly relevant to the answering process, facilitating a deep fusion of unstructured and structured knowledge and providing interpretability.

\subsubsection{Model Training} 
We use the cross-entropy function as the loss function, with the training objective of minimizing the cross-entropy loss between the ground-truth choice and the model's predicted choice.
\begin{equation}
{\mathcal L}\left(P^{*},P\right)=\sum_{i=1}^{|P|}p_{i}^{*}\log{(\frac{1}{p_{i}})}
\end{equation}

\section {Experiment}

\subsection{Knowledge Sources and Datasets}
We incorporate ConceptNet ~\cite{DBLP:conf/aaai/SpeerCH17}, Wikipedia ~\cite{wu2007autonomously} and Cambridge Dictionary ~\cite{matsumoto2009cambridge} as external knowledge sources. To evaluate G-SAP, we access its performance on three benchmark datasets: Commonsense-QA ~\cite{talmor2019commonsenseqa}, OpenbookQA ~\cite{mihaylovcan}, and PIQA \cite{bisk2020piqa}.

\noindent \textbf{CommonsenseQA ~\cite{talmor2019commonsenseqa}} is a 5-choices QA dataset, containing 12,102 questions. It requires conceptual commonsense knowledge for reasoning. We use the official split dataset.

\noindent \textbf{OpenBookQA ~\cite{mihaylovcan}} is a 4-choices QA dataset of 5,957 questions about elementary scientific knowledge; splits by ~\cite{mihaylovcan}.

\noindent \textbf{PIQA ~\cite{bisk2020piqa}} is a 2-choices QA dataset about physics commonsense knowledge. We conduct experiments on the development set (Dev) as the test set is not public.

\subsection{Implementation Details}
Given a question and options, we extract up to 100 two-hop knowledge paths from ConceptNet based on the key entities in the question and options. We retrieve the top 10 Wikipedia evidence from an Elasticsearch engine and use Semantic Role Labeling (SRL) to extract triples (subject, predicate, object). We also retrieve paraphrases of key entities from the Cambridge Dictionary and align all the evidence to generate an evidence graph. In our experimental setup, we employ several pre-trained language models (PLMs) including RoBERTa-large ~\cite{liu2019roberta}, AristoRoBERTa ~\cite{clark2020f}, or RoBERTa-xlarge ~\cite{liu2019roberta}. We utilize Adamw as the optimizer. For the specific hyperparameters, we configure the learning rate of the language model to $1e-5$ and the learning rate of the graph network to $1e-4$. The feature dimension of the graph network is set to 300, and we employ 5 layers in the graph network. Furthermore, we use a batch size of 1 and a maximum sequence length of 256. To ensure a stable training process, we perform warm-up update steps for 600 iterations and accumulated gradients for 1 step. Finally, we conducted 6 training rounds. The experiments are executed on a setup with Intel (R) Xeon (R) Platinum 8462Y+ CPU, $8$ NVIDIA A$800$-SXM$4$-$80$GB GPUs, and $1$TB RAM.

\subsection{Baselines}
\begin{spacing}{0.94}
Initially, we take \textbf{Random} ~\cite{wang2023car} and \textbf{Majority} ~\cite{wang2023car} as the baselines. The random method randomly takes a label as the answer, while the majority method takes the most frequent label as the answer. Subsequently, we proceed to choose additional baselines as below:

\noindent \textbf{Language models} include RoBERTa-large ~\cite{liu2019roberta}, BERT-base ~\cite{kenton2019bert}, GPT2-large ~\cite{radford2019language}, GPT-3.5 ~\cite{brown2020language}  which are fine-tuned on training data and making predictions on test data without external knowledge.

\noindent \textbf{KagNet} ~\cite{lin2019kagnet} converts question-answer pairs from the semantic space to a pattern graph in the knowledge-based symbolic space, i.e., relevant subgraphs of an external knowledge graph, for interpretable reasoning.

\noindent \textbf{XLNET+Graph Reasoning} ~\cite{lv2020graph} constructs evidence knowledge subgraphs from ConceptNet and Wikipedia. It proposes a graph-based approach, including a contextual representation learning module and a reasoning module.

\noindent \textbf{ALBERT+Path Generator} ~\cite{wangconnecting} proposes a multi-hop knowledge path generator to dynamically generate evidence by utilizing ConceptNet. It incorporates a PLM to exploit the vast amount of unstructured knowledge available to compensate for the incompleteness of the knowledge base.

\noindent \textbf{ALBERT+KD} ~\cite{DBLP:conf/coling/ChenJCZ20} proposes a graph-based iterative knowledge retrieval method, which ranks the importance of the retrieved knowledge from multiple knowledge sources and uses an answer selection attention module for reasoning.
\end{spacing}

\noindent \textbf{FeQA} ~\cite{zhang2023feqa} is a QA system model that leverages large-scale pre-trained language models and knowledge graphs. It incorporates a dual-attention mechanism that utilizes sources like Wiktionary and other QA datasets to enhance the semantic comprehension of the questions and integrates GNNs to facilitate entity inference.

Additionally, to ensure fairness, we also conduct experiments only utilizing the LM+KG method. We use RoBERTa-large ~\cite{liu2019roberta} as the backbone, the other baselines include Fine-tuned RoBERTa-large, RN ~\cite{santoro2017simple}, KagNet ~\cite{lin2019kagnet}, RGCN ~\cite{vanmodeling}, GconAttn ~\cite{wang2019improving}, MHGRN ~\cite{feng2020scalable}, QA-GNN ~\cite{yasunaga2021qa}, MVP-Tuning ~\cite{huang2023mvp}. 

\subsection{Result and Analysis}
\begin{spacing}{0.91}

\begin{table*} [!t]\tiny
\caption{Baseline Model Comparison Result}

\label{label1}
\centering
\scalebox{0.75}{
\resizebox{\linewidth}{!}{
\begin{tabular}{c c c c c c}
\hline
\multirow{2}{*}{Methods}   & \multicolumn{2}{c}{OpenbookQA}         & \multicolumn{2}{c}{CSQA}     & PIQA    \\ \cline{2-6} 
                          & \multicolumn{1}{c}{Dev Acc} & Test Acc & Dev Acc      & Test Acc & Dev Acc \\ \hline
Random                    & \multicolumn{1}{c}{-}       & 25.00     & -            & 20.00     & 50.00    \\ 
Majority                  & \multicolumn{1}{c}{-}       &   25.70       & -            & 20.90     & 50.50    \\ 
\hline
RoBERTa-large ~\cite{liu2019roberta}            & \multicolumn{1}{c}{63.72}   & 59.62    & 51.23        & 45.00      & 67.61    \\
BERT-base ~\cite{kenton2019bert}            & \multicolumn{1}{c}{65.53}   & 57.62     & 58.38        & 53.08    & 64.86\\
GPT2-Large ~\cite{radford2019language}                & \multicolumn{1}{c}{-}       & 53.21     & -            & 41.39     & 68.88    \\ 
GPT-3.5 (text-davinci-003) ~\cite{brown2020language} & \multicolumn{1}{c}{-}       & 71.51     & -            & 68.92     & 67.80    \\ 
GPT-3.5 (gpt-3.5-turbo) ~\cite{brown2020language}   & \multicolumn{1}{c}{-}       & 72.90     & -            & 74.48     & 75.11    \\
\hline
KagNet (BERT) ~\cite{lin2019kagnet}              & \multicolumn{1}{c}{67.81}    & 62.33    & 64.46        & 58.90     & 72.44   \\  
XLNET+Graph Reasoning ~\cite{lv2020graph}     & \multicolumn{1}{c}{81.58}    & 77.50     & 79.31        & 75.29     & 80.47   \\ 
ALBERT+Path Generator ~\cite{wangconnecting}     & \multicolumn{1}{c}{78.91}    & 76.79     & 78.42        & 76.12     & 80.57   \\ 
ALBERT+KD ~\cite{DBLP:conf/coling/ChenJCZ20}                 & \multicolumn{1}{c}{82.21}    & 80.88     & 82.39        & 77.32     & 81.34        \\ 
FeQA (RoBERTa-large) ~\cite{zhang2023feqa}               & \multicolumn{1}{c}{72.58}    & 70.41     & 79.81        & 76.23     & 74.32   \\ 
\hline
G-SAP (RoBERTa-large)         & 80.83  & 76.84    & 81.58       & 78.12    & 81.42   \\
G-SAP (AristoRoBERTa)         & \multicolumn{1}{c}{\textbf{90.98}}   & \textbf{87.65}    & \textbf{87.45}        & \textbf{83.72}    & \textbf{90.49}   \\ 
G-SAP (RoBERTa-xlarge)        & \multicolumn{1}{c}{\textbf{\underline{91.35}}}   & \textbf{\underline{87.83}}    & \textbf{\underline{87.96}}        & \textbf{\underline{84.52}}    & \textbf{\underline{90.97}}   \\ 
\hline
Human Performance         & \multicolumn{1}{c}{-}       & 91.70     & \multicolumn{1}{c}{-}            & 88.90     & 94.90    \\ \hline
\end{tabular}
}
}

\end{table*}

\textbf{Baseline Comparison:} Table \ref{label1} presents a comprehensive comparison of our proposed G-SAP with popular methods on the OpenbookQA, CSQA, and PIQA datasets. For PIQA, we report the comparison results on the development set since the annotations of the PIQA test set have not been released. G-SAP attains superior performance across all baseline methods on the three datasets. Notably, the G-SAP with pre-trained language model RoBERTa-xlarge outperforms the previous outstanding method ALBERT+KD by approximately 6.95\% on the OpenbookQA test set, 7.20\% on the CSQA test set, and 9.63\% on the PIQA development set. The 7.93\% average improvement highlights the powerful cross-modal ability present in our model compared to others.
\begin{table}[h]
\small
\caption{LM+GNN models Comparison Result}
\centering
\label{gnn}
\resizebox{\linewidth}{!}{%
\begin{tabular}{c c c c c c}
\hline
&\multicolumn{2}{c}{OpenbookQA}             &\multicolumn{2}{c}{CSQA}                   & PIQA                                         \\ \cline{2-6} 
\multirow{2}{*}[3ex]{\centering Methods}    & Dev Acc                 & Test Acc                             & Dev Acc                                     & Test Acc               &Dev Acc
\\ \hline\\
\multirow{2}{*}[3ex]{\makecell{Fine-Tuned \\ PLM (w/o KG)}} & {\color[HTML]{333333} \multirow{2}{*}[3ex]{\centering\textbf{70.22}}} & {\color[HTML]{333333} \multirow{2}{*}[3ex]{\centering\textbf{64.80}}} & {\color[HTML]{333333} \multirow{2}{*}[3ex]{\centering\textbf{71.51}}} & {\color[HTML]{333333} \multirow{2}{*}[3ex]{\centering\textbf{66.24}}}& {\color[HTML]{333333} \multirow{2}{*}[3ex]{\centering\textbf{70.03}}} \\ \hline
+RN ~\cite{santoro2017simple}                              & {\color[HTML]{333333} 70.87} & {\color[HTML]{333333} 65.20}  & {\color[HTML]{333333} 74.57} & {\color[HTML]{333333} 69.08} & {\color[HTML]{333333} 74.39}\\
+KagNet ~\cite{lin2019kagnet}  & {\color[HTML]{333333} 72.50} & {\color[HTML]{333333} 70.67} & {\color[HTML]{333333} 71.76} & 
{\color[HTML]{333333} 68.59} & {\color[HTML]{333333} 71.65}\\
+RGCN ~\cite{vanmodeling}               & {\color[HTML]{333333} 67.12} & {\color[HTML]{333333} 62.45} & {\color[HTML]{333333} 72.69} & {\color[HTML]{333333} 68.41} & {\color[HTML]{333333} 71.97}\\
+GconAttn ~\cite{wang2019improving}  & {\color[HTML]{333333} 70.39} & {\color[HTML]{333333} 64.95} & {\color[HTML]{333333} 72.61} & {\color[HTML]{333333} 68.59} & {\color[HTML]{333333} 73.64}\\  
+MHGRN ~\cite{feng2020scalable}                           & {\color[HTML]{333333} 73.33} & {\color[HTML]{333333} 66.85} & {\color[HTML]{333333} 74.45}  & {\color[HTML]{333333} 71.11} & {\color[HTML]{333333} 74.13}\\
+QA-GNN ~\cite{yasunaga2021qa}         & {\color[HTML]{333333} 73.54} & {\color[HTML]{333333} 70.58} & {\color[HTML]{333333} 76.54} & {\color[HTML]{333333} 73.41}& {\color[HTML]{333333} 76.29} \\
+MVP-Tuning ~\cite{huang2023mvp}         & {\color[HTML]{333333} 77.68} & {\color[HTML]{333333} 70.72} & {\color[HTML]{333333} 80.16} & {\color[HTML]{333333} 74.35}& {\color[HTML]{333333} 77.32} \\
\hline
\multirow{2}{*}[1.1ex]{\centering G-SAP}      & \multicolumn{1}{c}{\textbf{80.83}}       & \textbf{76.84}    & \textbf{81.58}    & \textbf{78.12}  & \textbf{81.42} \\ 
\hline
\end{tabular}%
}

\end{table}

To ensure a more fair comparison, we also evaluate LM+GNN models. In Table \ref{gnn}, the results demonstrate that G-SAP consistently outperforms all baselines. Notably, it has superior performance compared to the second-best mothed, MVP-Tuning ~\cite{huang2023mvp}, with a margin ranging from $1.42$\% to $6.12$\%, and an average improvement of  $3.71$\%.
By incorporating structure-aware prompt learning with a graph-based reasoning model, we achieve a deep interaction of heterogeneous knowledge, effectively enhancing CSQA performance.

\subsection{Ablation Experiment}

In ablation experiments, We primarily utilize the CSQA dataset to evaluate the performance of different variants.

\textbf{The impact of PLM module:} To validate the importance of the pre-training module in our G-SAP, we compared it with the following variants: (1) G-SAP-w/o-SAPL: A variant of G-SAP without SAPL module. (2) G-SAP-w/o-Prompt: A variant of G-SAP without all the prompts in the PLM. (3) G-SAP-w/o-Pa nodes: Remove all the nodes and relations generated by the paraphrase in the evidence subgraph. (4) G-SAP-w/o-Pa texts: Remove all the paraphrased texts in PLM. (5) G-SAP-w/o-Pa: A variant of G-SAP obtained by removing both paraphrase nodes and texts.

\begin{table}[h]\small
\caption{The impact of PLM Module}
\centering
\label{ab1}
\scalebox{0.7}{
\resizebox{\columnwidth}{!}{%
\begin{tabular}{c c c}
\hline
Methods                                    & CSQA Dev                      & CSQA Test                         \\ \hline   
G-SAP-w/o-SAPL                             & {\color[HTML]{333333} 66.76} & {\color[HTML]{333333} 64.81} \\
G-SAP-w/o-Prompt                             & {\color[HTML]{333333} 84.65} & {\color[HTML]{333333} 81.53} \\
G-SAP-w/o-Pa nodes                        & {\color[HTML]{333333} 86.89} & {\color[HTML]{333333} 83.32} \\ 
G-SAP-w/o-Pa texts                          & {\color[HTML]{333333} 87.06} & {\color[HTML]{333333} 83.78} \\ 
G-SAP-w/o-Pa                               & {\color[HTML]{333333} 86.89} & {\color[HTML]{333333} 82.99} \\ 
\hline
G-SAP                                     & {\color[HTML]{333333} \textbf{87.96}} & {\color[HTML]{333333} \textbf{84.52}} \\ \hline
\end{tabular}%
}
}
\end{table}
Table \ref{ab1} shows the results of the ablation study. It is observed that removing only prompts in the PLM restricts the model's capability to handle complex tasks like commonsense question answering, particularly when a large number of parameters of PLM are fixed. The observed drop in performance when excluding the entire pre-trained language model (PLM) highlights the significance of fully interacting and jointly encoding both structured and unstructured knowledge. Additionally, relying solely on the question and choice text is insufficient, and the incorporation of paraphrases and paraphrase entities can significantly enhance the model's ability

\textbf{The impact of Graph-based module:} To validate the importance of our G-SAP graph-based reasoning module, we conduct a comparison with the following variants: (1) G-SAP-w/o-HMPR: A variant of G-SAP by removing HMPR module. (2) G-SAP-w/o-BiGRU: A variant of G-SAP obtained by deleting the BiGRU component. (3) G-SAP-w/o- Relevance Score: A variant of G-SAP without using relevance scores. (4) G-SAP-w/o-Graph Attention Weights: A variant of G-SAP by removing attention-based weight acquisition.

\begin{table}[h]\small
\caption{The impact of Graph-based module}

\centering
\label{ab2}
\scalebox{0.9}{
\resizebox{\columnwidth}{!}{%
\begin{tabular}{c c c}
\hline

Methods                               & CSQA Dev                        & CSQA Test                         \\ \hline 
G-SAP-w/o-HMPR module                        & {\color[HTML]{333333}  79.15    } & {\color[HTML]{333333}  76.01    } \\  
G-SAP-w/o-BiGRU                        & {\color[HTML]{333333}   87.43   } & {\color[HTML]{333333} 83.07     } \\ 
G-SAP-w/o-Relavance Score              & {\color[HTML]{333333} 87.21} & {\color[HTML]{333333} 83.63} \\ 
G-SAP-w/o-Graph Attention Weights                & {\color[HTML]{333333} 87.51} & {\color[HTML]{333333} 83.85} \\ 
\hline
G-SAP                                & {\color[HTML]{333333} \textbf{87.96}} & {\color[HTML]{333333} \textbf{84.52}} \\ \hline
\end{tabular}%
}
}

\end{table}

As shown in Table \ref{ab2}, it is evident that the performance considerably declines when the graph reasoning component is omitted. This observation further highlights the necessity and effectiveness of graph-based integrating structured and unstructured knowledge. The presence of the BiGRU component ensures accurate final answers by enabling effective long-sequence reasoning. Additionally, the attention weights allocate appropriate weights based on variable similarities, leading to more reasonable computations.

\textbf{The impact of Prompt:} To verify the effectiveness of the prompt, we selected the following variations for experimentation. (1) G-SAP-w/o-Prompt-entity: A variant of G-SAP without the entity prompt $F(E^h_i,g)$ and $F(E^t_i,g)$. (2) G-SAP-w/o-Prompt-relation: A variant of G-SAP without the relation prompt $F(R_i,g)$. (3) G-SAP-w/o-Prompt: A variant of G-SAP without the whole structure-aware prompt $F(E^h_i,g)$, $F(R_i,g)$ and $F(E^t_i,g)$. (4) G-SAP-w/o-Random Prompt: A variant of G-SAP using a random prompt.
\begin{table}[!h]\tiny
\caption{Impact of Prompt}
\centering
\label{ab3}
\scalebox{0.9}{
\resizebox{\columnwidth}{!}{%
\begin{tabular}{c c c}
\hline
Methods                                    & CSQA Dev                      & CSQA Test                         \\ \hline 
G-SAP-w/o-Prompt-entity                   & {\color[HTML]{333333} 86.26} & {\color[HTML]{333333} 83.63} \\  
G-SAP-w/o-Prompt-relation                 & {\color[HTML]{333333} 85.31} & {\color[HTML]{333333} 82.04} \\ 
G-SAP-w/o-Prompt                          & {\color[HTML]{333333} 84.65} & {\color[HTML]{333333} 81.53} \\ 
G-SAP-w-Random Prompt                       & {\color[HTML]{333333} 84.96} & {\color[HTML]{333333} 81.89} \\ 
\hline
G-SAP                                     & {\color[HTML]{333333} \textbf{87.96}} & {\color[HTML]{333333} \textbf{84.52}} \\ \hline
\end{tabular}%
}
}
\end{table}

The results in Table \ref{ab3} show that removing any part of the prompt or using a random prompt leads to a partial decrease in performance. Furthermore, when the prompt is completely removed, the model struggles to adapt to downstream QA tasks with fixed PLM parameters. This finding provides strong evidence for the effectiveness of incorporating prompts in the G-SAP model.

\begin{figure}[!t]
    \centering
    \includegraphics[scale=0.35]{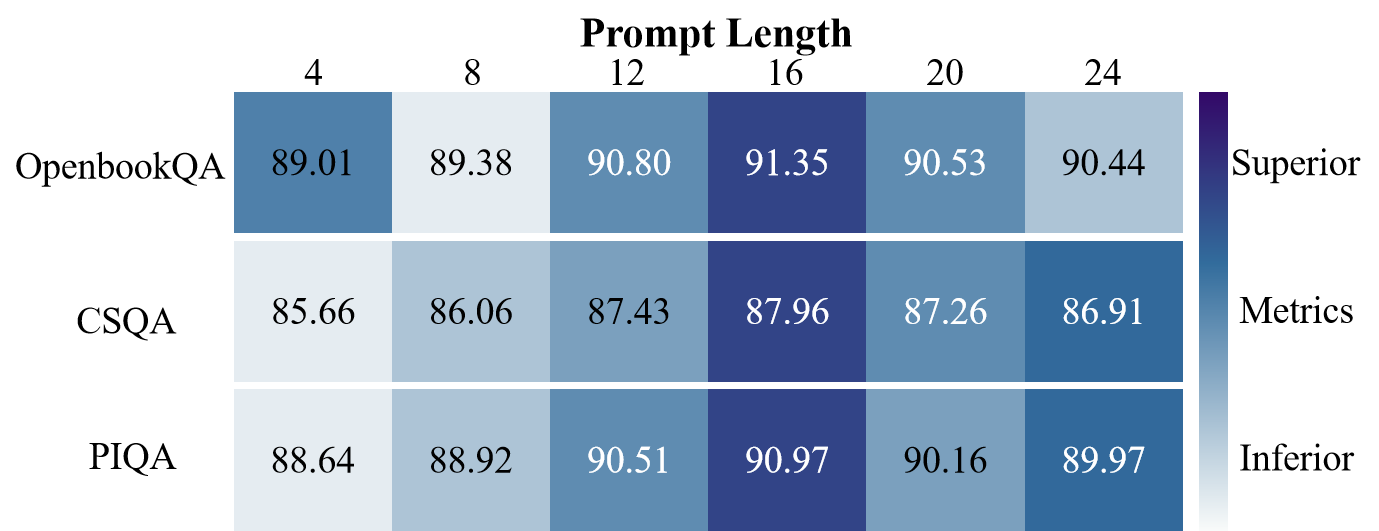}
    \caption{The impact of prompt length
    }
    \label{fig:prompt}

\end{figure}

Similarly, we conduct extensive experiments to assess the impact of prompt length in G-SAP. As depicted in Figure \ref{fig:prompt}, the results reveal a notable improvement in model performance with increasing prompt length, up until 16. The model with a prompt length of 16 outperformed the worst prompt length by 2.32\% on average. However, beyond this point, there is no significant gain observed, suggesting the existence of an optimal balance point.

\textbf{The impact of KG source:} 
To investigate the impact of different knowledge sources on our work, we conduct experiments by building variants based on different KG sources. (1) G-SAP (w/o KG): A variant of G-SAP without all external knowledge. (2) +ConceptNet: A variant of G-SAP only using conceptNet. (3) +Wikipedia: A variant of G-SAP only using wikipedia. (4) +Cambridge Dictionary: A variant of G-SAP only using Cambridge Dictionary. (5) +Concept+Wikipedia: A variant of G-SAP both using ConceptNet and Wikipedia.  
(6) +ALL: A variant of G-SAP using ConceptNet, Wikipedia, and Cambridge Dictionary simultaneously.
\begin{table}[!h]\tiny
\caption{Knowledge Source Ablation}
\centering
\label{ab4}
\scalebox{0.88}{
\resizebox{\columnwidth}{!}{%
\begin{tabular}{c c c}
\hline
Methods (knowledge sources)               & CSQA Dev                       & CSQA Test\\ \hline 
G-SAP (w/o KG)                           & {\color[HTML]{333333} 76.45} & {\color[HTML]{333333} 72.17}\\ \hline 
+ConceptNet only                         & {\color[HTML]{333333} 83.48} & {\color[HTML]{333333} 79.04}\\ 
+Wikipedia only                          & {\color[HTML]{333333} 81.50} & {\color[HTML]{333333} 77.64}\\ 
+Cambridge Dictionary only                    & {\color[HTML]{333333} 78.51} & {\color[HTML]{333333} 72.41}\\ 
+ConceptNet+Wikipedia                      & {\color[HTML]{333333} 85.56} & {\color[HTML]{333333}81.42}\\ \hline
+ALL                                     & {\color[HTML]{333333} \textbf{87.96}} & {\color[HTML]{333333} \textbf{84.52}}\\ \hline
\end{tabular}%
}
}
\end{table}

The results presented in Table \ref{ab4} indicate that the absence of external knowledge sources significantly diminishes the model's reasoning capability. Given the challenging nature of commonsense question answering, PLMs rely on rich external knowledge to accurately reason. Our model leverages the combined knowledge from ConceptNet, Wikipedia, and the Cambridge Dictionary to achieve optimal performance.

\section{Conclusions}
In this paper, we propose G-SAP, a graph-based structure-aware prompt learning model for commonsense reasoning, which effectively tackles the over-fitting issue toward textual information and facilitates deep interaction among heterogeneous modalities of knowledge. It comprises three primary modules: an evidence graph generation (EGG) module, a structure-aware prompt learning (SAPL) module, and a heterogeneous message passing reasoning (HMPR) module. EGG attempts to generate comprehensive evidence graphs from multiple sources. SAPL is the key innovation of G-SAP that bridges a graph-based message-passing model and a frozen PLM through the structure-aware prompts generated from the representations of graph entities and relations, thereby avoiding the textual over-fitting issue. To facilitate a deep interaction of heterogeneous knowledge,  HMPR introduces graph-based networks and attention mechanisms to respectively
fuse and reason over the different modalities. Extensive results on three benchmark datasets demonstrate the superiority of our proposed model. In the future, we will focus on more KG question-answering scenarios, e.g., the open-domain question-answering tasks that require external background knowledge.
\end{spacing}
\bibliographystyle{ACM-Reference-Format}

\bibliography{G-SAP}

\end{sloppypar}
\end{document}